\documentclass{article}
\usepackage{spconf}

\usepackage{amsmath,graphicx,bbm,booktabs,microtype,balance}
\usepackage[hidelinks]{hyperref}

\newcommand{\etal}{\textit{et al}.}
\newcommand{\ie}{\textit{i}.\textit{e}.}

\DeclareMathOperator*{\argmax}{arg\,max}

\DeclareMathOperator*{\similarity}{sim}

\title{A Location-Sensitive Local Prototype Network\\ for Few-Shot Medical Image Segmentation \\[-1.0ex]}

\name{Qinji Yu\textsuperscript{1,*}, Kang Dang\textsuperscript{2,*}, Nima Tajbakhsh\textsuperscript{2}, Demetri Terzopoulos\textsuperscript{3,2}, Xiaowei Ding\textsuperscript{1,2}  \thanks{*\,Contributed equally.} }
\address{\textsuperscript{1}\,Shanghai Jiao Tong University, China\\ \textsuperscript{2}\,VoxelCloud, Inc., Los Angeles, USA\\ \textsuperscript{3}\,University of California, Los Angeles, USA\\[-2.0ex]}

\begin{document}

\maketitle

\begin{abstract}
Despite the tremendous success of deep neural networks in medical image segmentation, they typically require a large amount of costly, expert-level annotated data. Few-shot segmentation approaches address this issue by learning to transfer knowledge from limited quantities of labeled examples. Incorporating appropriate prior knowledge is critical in designing high-performance few-shot segmentation algorithms. Since strong spatial priors exist in many medical imaging modalities, we propose a prototype-based method---namely, the location-sensitive local prototype network---that leverages spatial priors to perform few-shot medical image segmentation. Our approach divides the difficult problem of segmenting the entire image with global prototypes into easily solvable subproblems of local region segmentation with local prototypes. For organ segmentation experiments on the VISCERAL CT image dataset, our method outperforms the state-of-the-art approaches by 10\% in the mean Dice coefficient. Extensive ablation studies demonstrate the substantial benefits of incorporating spatial information and confirm the effectiveness of our approach. 
\end{abstract}


\section{Introduction}
\label{sec:intro}

The inception of deep neural networks has revolutionized the landscape of medical image segmentation~\cite{ronneberger2015u, isensee2018nnu, chen2017rethinking}. Despite their tremendous success, however, depriving them of sufficient quantities of labeled images limits their effectiveness.  Unfortunately, obtaining abundant amounts of expert-level-accurate, pixel-wise annotated data in the medical imaging domain incurs significant time and expense~\cite{tajbakhsh2019surrogate}. Therefore, it is imperative to explore new medical image segmentation techniques that can learn from scarcely annotated data.

Few-shot semantic segmentation is a relatively new trend in computer vision that addresses the data scarcity issue \cite{rakelly2018few, DBLP:conf/bmvc/DongX18,wang2019panet, zhang2019canet, liu2020prototype, roy2020squeeze}. A few-shot segmentation system learns how to transfer knowledge from a few labeled samples, typically referred to as support images, to guide image segmentation with novel classes, known as queries.  The concept is motivated by human learning---the process of rapidly learning new knowledge from a few observations by leveraging strong prior knowledge acquired from experience. Analogously, incorporating appropriate prior knowledge is critical in the development of high-performance few-shot semantic segmentation algorithms. It is well-known that strong spatial priors are inherent in many medical imaging modalities, including abdominal CT scans~\cite{dang2017learning, haghighi2020learning}.  For instance, the liver typically appears on the left side of the CT image slice, while the spleen appears on the right. Such spatial layout priors convey crucial information in the pixel-wise labeling of certain organ classes. This motivates us to study how to exploit spatial layout information within a few-shot image segmentation algorithm. To our knowledge, ours is the first attempt at incorporating spatial layout information in a few-shot medical image segmentation application.

\begin{figure*}
\centering
\includegraphics[width=0.8\linewidth]{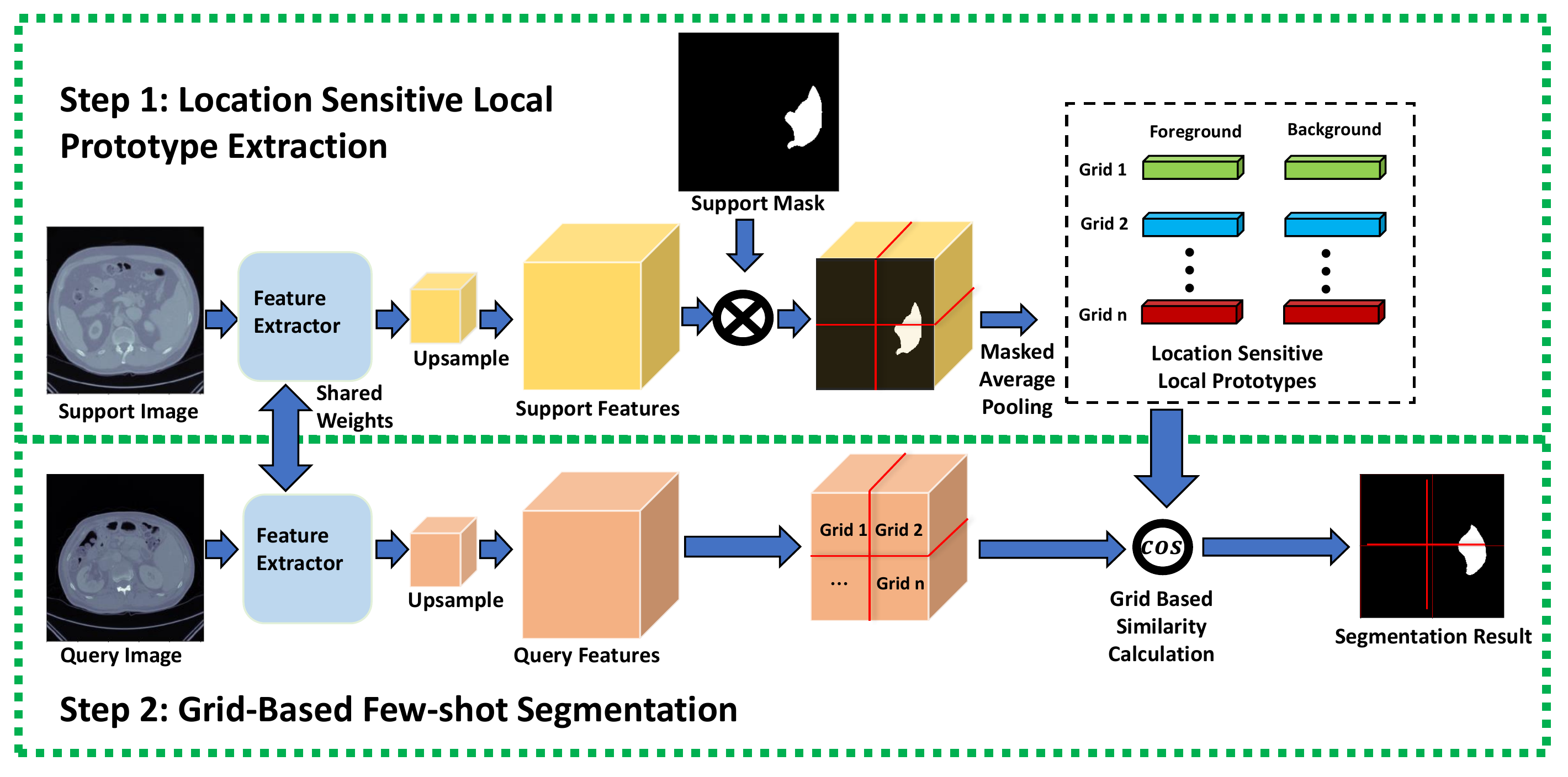}
\caption{Processing pipeline of our location-sensitive local prototype network.}
\label{fig:pipeline}
\end{figure*}

In particular, our approach is a prototype-based method~\cite{wang2019panet, zhang2019canet} that we call the \emph{location-sensitive local prototype network}.  We divide the image into overlapping image grids and extract location-specific local prototypes for all the support image grids. Unlike the global prototypes used in the aforecited publications, our local prototype is constrained to utilize the information within the corresponding grid; hence, it is location sensitive. Assuming that the support and query images have a similar spatial layout, each location in the query image can be matched to the corresponding local prototypes for pixel-wise segmentation. In this way, we divide the difficult problem of segmenting the entire image with global prototypes into easily solvable subproblems of local region segmentation with local prototypes, which leads to significantly improved performance. Hence, our contributions are as follows:
\begin{itemize}
\setlength\itemsep{-1pt}
\item Our novel location-sensitive local prototype network incorporates spatial layout information into the few-shot medical image segmentation application and demonstrates the substantial benefits of incorporating such information. 
\item We thoroughly analyze the grid size and other critical parameters in our experiments.
\item We establish a new state-of-the-art performance for organ segmentation on the VISCERAL contrast-enhanced CT image dataset~\cite{jimenez2016cloud} by showing that our method outperforms the currently best approach~\cite{roy2020squeeze} by 10\% in the mean Dice coefficient, thereby confirming the effectiveness of our proposed approach.
\end{itemize}

\section{Related Work}
\label{sec:format}

Few-shot semantic segmentation is a recent research topic. Promising schemes have appeared in recent years \cite{rakelly2018few, DBLP:conf/bmvc/DongX18, wang2019panet, zhang2019canet, liu2020prototype, roy2020squeeze}. Some efforts have focused on the few-shot medical image segmentation application~\cite{DBLP:journals/corr/abs-1810-12241, zhao2019data, roy2020squeeze, ouyang2020self}. Tajbakhsh \etal~\cite{tajbakhsh2020embracing} have surveyed this topic. Roy \etal~\cite{roy2020squeeze} adopted squeeze-excitation modules to effectuate the interaction between support and query images in order to perform few-shot organ segmentation. Many approaches are based on non-parametric metric learning frameworks, such as prototype networks~\cite{DBLP:conf/bmvc/DongX18, wang2019panet, liu2020prototype}. Wang \etal~\cite{wang2019panet} proposed a prototype alignment regularization strategy between the support and query to learn a better class-specific global prototype representation. Ouyang \etal~\cite{ouyang2020self} designed an adaptive local prototype pooling module for prototypical networks to solve the foreground-background imbalance problem in medical image segmentation. Our approach is also prototype-based; however, unlike previous work, we focus on incorporating spatial layout information and demonstrate it to be highly beneficial to few-shot medical image segmentation.

\section{Methodology}

We first introduce the few-shot segmentation problem in Sec.~\ref{sec:formulation}, then explain the two main steps of our location-sensitive local prototype network (Fig.~\ref{fig:pipeline}): location-sensitive local prototype extraction in Sec.~\ref{sec:prototype_extraction}, followed by grid-based few-shot segmentation in Sec.~\ref{sec:grid_based_seg}.

\subsection{Problem Setting}
\label{sec:formulation}

The problem is to train a model on a large labeled training dataset ${\cal D}_\text{train}$ that can perform segmentation on a testing dataset ${\cal D}_\text{test}$ with only a few annotated examples, where the class set ${\cal C}_\text{train} \in {\cal D}_\text{train}$ has no overlap with the class set ${\cal C}_\text{test} \in {\cal D}_\text{test}$; \ie, ${\cal C}_\text{train} \cap {\cal C}_\text{test} = \emptyset$. For example, given the training set with training classes ${\cal C}_\text{train}=\{\texttt{lungs}, \texttt{heart}\}$, we expect the trained model to be proficient at segmenting a different set of testing classes ${\cal C}_\text{test}=\{\text{\texttt{kidney}, \texttt{bones}}\}$ with reference only to a few labeled images in ${\cal D}_\text{test}$.  

Like previous research~\cite{wang2019panet, ouyang2020self}, we adopt an episodic paradigm for the few-shot segmentation task. Given an $n$-way, $k$-shot task, during model training we sample a set of episodes ${\cal T}_\text{train}$ from ${\cal D}_\text{train}$. Each episode in ${\cal T}_\text{train}$ is composed of a support set ${\cal S}$ and a query set ${\cal Q}$, and the model is trained to transfer segmentation cues from ${\cal S}$ to ${\cal Q}$ in a class-agnostic manner. The support set ${\cal S}=\{(I_i^s,M_{i,c}^s)\}$ contains $1\le i \le k$ pairs of support images $I_i^s$ and masks $M_{i,c}^s$ corresponding to a set of $1\le c \le n$ randomly picked classes ${\cal C}\in{\cal C}_\text{train}$. The query set ${\cal Q}=\{(I_j^q,M_{j,c}^q)\}$ contains $1\le j \le n_q$ different query pairs likewise sampled from ${\cal D}_\text{train}$ with class set ${\cal C}$. As each episode randomly samples different semantic classes, the trained model is expected to be class-agnostic and generalize well.

After training the segmentation model, we evaluate its performance on ${\cal D}_\text{test}$. During model testing, a set of episodes ${\cal T}_\text{test}$ is obtained in similar fashion from ${\cal D}_\text{test}$, where the masks are associated with testing classes ${\cal C}_\text{test}$. For each testing episode, the segmentation model is evaluated on the query set ${\cal Q}$ given the support set ${\cal S}$.

\subsection{Location-Sensitive Local Prototype Extraction}
\label{sec:prototype_extraction}

Consider a coarse set ${\cal G}=\{g_m\}$ of $1\le m \le n_g$ overlapping image grids uniformly distributed across the $w\times h$ sized image. Grid $g_m$ of size $\alpha w \times \alpha h$, where $\alpha$ denotes the grid scale, is centered at $(x_m,y_m)$. To extract location-sensitive local prototypes, support images are passed to the feature encoder to generate the corresponding features maps. The location-sensitive local prototype of $g_m$ is calculated from support images and masks by masked average pooling, as follows:
\begin{equation}
p_{c,g_m}^s=\frac{1}{k}\sum_{i=1}^k\frac{\sum_{(x,y)\in g_m}{F_{i}^s(x,y) M_{i,c}^s(x,y)}}{\sum_{(x,y)\in g_m}M_{i,c}^s(x,y)},  
\end{equation}
where $(x,y)$ are spatial coordinates and $F_{i}^s$ are the extracted feature maps. If $\sum_{\left(x,y\right)\in g_m}M_{i,c}^s(x,y)=0$, we simply set $p_{c,g_m}^s=0$. We also treat the background as a semantic class whose prototype $p_{0,g_m}^s$ is computed by performing masked average pooling on all locations within $g_m$ excluding any foreground classes. The following are key observations regarding the extracted local prototype $p_{c,g_m}^{s}$:
\begin{enumerate}
\item Prototype $p_{c,g_m}^{s}$ is associated with the local image grid $g_m$ and, hence, is location sensitive. Grid scale $\alpha$ is a critical parameter influencing the spatial information encoded by the local prototype. When $\alpha$ is so small that the grid contains only a single pixel, the extracted local prototype becomes overly location sensitive, whereas when $\alpha=1$, the grid expands to the full size of the image and position information is not encoded at all. 
\item Unlike global prototypes, our local prototype $p_{c,g_m}^{s}$ is constrained to utilize information only within the corresponding grid $g_m$. Hence, the loss of local information due to the averaging out of features outside the local grid can be prevented. This is essential for constructing high-quality background class prototypes $p_{0,g_m}^{s}$, because the background class tends to be spatially inhomogeneous, involving many different anatomical structures whose information would be unreasonably averaged out in the absence of  the local grid constraint.
\end{enumerate}
We collect the extracted local prototypes for all the grid locations and classes into set ${\cal P}{=\{p_{c,g_m}^{s}\}}$, where $1\le m \le n_g$ and $0\le c \le n$. 

\subsection{Grid-Based Few-Shot Segmentation}
\label{sec:grid_based_seg}

Assuming that the support and query images have a similar spatial layout, local prototypes ${\cal P}$ extracted from the support set can be matched to the corresponding locations in the query images in order to perform grid-based few-shot segmentation. Given the support prototype set ${\cal P}$ of $n$ foreground classes and the background class, the probability score of the query image corresponding to class $c$ is obtained as
\begin{equation}
P_{j,c}^q(x,y) =\sigma\left(\similarity(F_{j}^q(x,y), {\cal P})\right),
\label{eq:global_prob}
\end{equation}
where $\similarity(F_{j}^q(x,y), {\cal P})$ is the similarity score between the query feature vector $F_{j}^q(x,y)$ and the support prototype set ${\cal P}$, and $\sigma$ is the softmax operation. 

For non-overlapping image grids, we may employ the cosine similarity between the query feature vector $F_{j}^q(x,y)$ and the corresponding matched grid prototype $p_{c,g_m}^s$, where the query feature vector location $(x,y)$ belongs to the grid $g_m$. With overlapping image grids, however, the feature at location $(x,y)$ may be mapped to multiple grids. We collect all the grids that influence location $(x,y)$ into a set $\Omega=\{g_m: (x,y)\in (x_m,y_m,\alpha)\}$ and calculate the similarity score as
\begin{equation}
\similarity(F_{j}^q(x,y), {\cal P})=\max_{g_m\in\Omega}\left(\cos(F_j^q(x,y), p_{c,g_m}^{s})\right), 
\end{equation}
where $\cos$ denotes the cosine similarity function. 

The network can be trained end-to-end using the standard cross-entropy loss: $L_\text{ce}(P^q, M^q)$, where $P^q$ and $M^q$ refer, respectively, to the predicted probability maps and ground-truth query masks of the particular training batch. For network evaluation, the predicted segmentation label of query image $I_j^q$ at spatial coordinate $(x,y)$ is then given by: $\hat{M}_j^q(x,y) =\argmax_{c} P_{j,c}^q(x,y)$.

\section{Experiments}

\begin{table}[t]
\small
\setlength{\tabcolsep}{0.5 mm}
\label{tab:my_label}
\centering
\begin{tabular}{lccccc}
\toprule 
Method&	\texttt{liver}&	\texttt{spleen}& \texttt{kidney}& \texttt{psoas}& Mean\\
\midrule 
PANet \cite{wang2019panet}&	59.4&	24.1&	23.7&	15.6&	30.7\\
SENet \cite{roy2020squeeze}& 70.0&	60.7&	46.4&	49.9&	56.7\\
\midrule
Our Result&	77.9&	71.5&	67.5&	49.9&	66.7\\
w/ Added Classes&	\textbf{79.3}&	\textbf{73.3}&	\textbf{76.5}&	\textbf{52.4}&	\textbf{70.3}\\
\midrule
ALPNet \cite{ouyang2020self}&	78.3&	70.9&	72.1&	-&	-\\
\bottomrule 
\end{tabular}
\caption{Comparisons against state-of-the-art methods.}
\label{table:sota}
\end{table}


For all our experiments, we utilized contrast-enhanced CT scans from the VISCERAL dataset~\cite{jimenez2016cloud}. We used 65 silver corpus scans for training and 20 gold corpus scans for testing. We performed 1-way, 1-shot experiments for six organ classes: \texttt{liver}, \texttt{spleen}, \texttt{left kidney}, \texttt{right kidney}, \texttt{left psoas}, \texttt{right psoas}. For a fair comparison, most experiments reported in Sec.~\ref{sec:sota} were performed using the same dataset configuration used by Roy \etal~\cite{roy2020squeeze}. In addition to the six organ classes, the VISCERAL dataset contains for each scan 14 additional organ annotations not used by these authors, which we use in other experiments. To evaluate 2D segmentation performance on 3D volumetric images, we follow the protocol described in~\cite{roy2020squeeze} and use the Dice score as the metric. 	 
	
Like~\cite{zhang2019canet}, we use as a feature encoder a customized version of ResNet-50~\cite{he2016deep} pre-trained on ImageNet~\cite{deng2009imagenet}. CT slices are resized to $512\times512$ before they are input to our network. We set the grid scale $\alpha=1/8$ and spacing between grid centers to half the grid size for most experiments. We apply RandomGamma, RandomContrast, and RandomBrightness to augment the training images. The models are trained by SGD~\cite{bottou2010large} with a batch size of 4 for 10,000 iterations. The initial learning rate is $10^{-3}$ and is decayed by 0.1 per 2,500 iterations. 

\subsection{Comparisons Against the State of the Art}
\label{sec:sota}

\begin{figure}
\centering
\includegraphics[width=\linewidth]{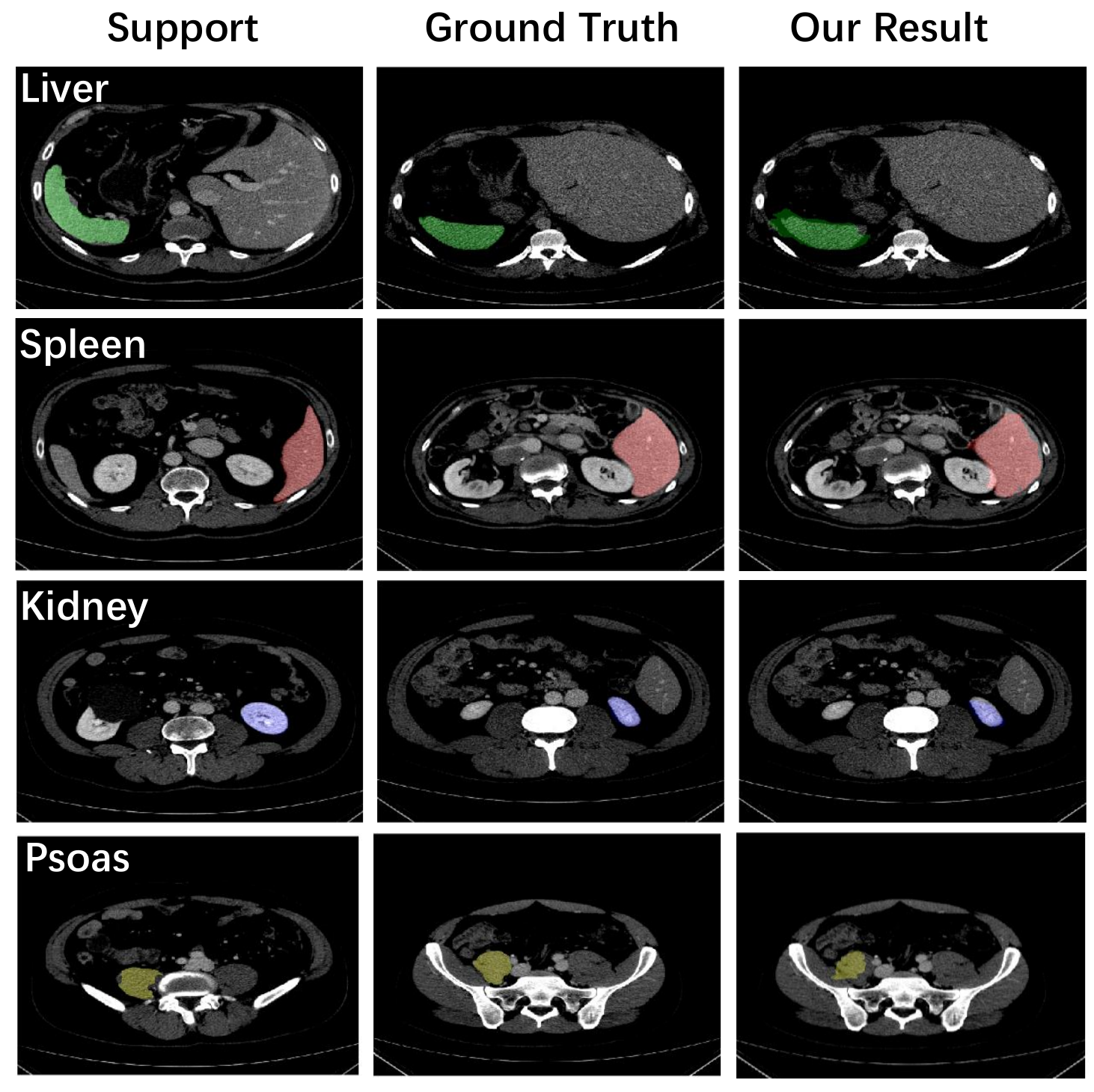}
\caption{Visual results. (Left) Support images and the associated ground-truth masks. (Middle) Query images and the associated ground-truth masks. (Right) Query images overlayed with our segmentation results.}
\label{fig:visual}
\end{figure}

Referring to Table~\ref{table:sota}, with an experimental protocol identical to that of the SENet few-shot organ segmentation model~\cite{roy2020squeeze}, our model achieved state-of-the-art performance in all the organ classes and a mean Dice coefficient of 66.7\%, compared to SENet's best-reported performance of 56.7\%. PANet~\cite{wang2019panet} is a popular prototype network previously tested only on natural image datasets; we experimented with its code on the VISCERAL dataset. Unlike our approach, both SENet and PANet do not explicitly incorporate a spatial layout prior. As is demonstrated by our ablation study, reported below, appropriately encoding spatial information is critical for good performance, and explains why our method performs significantly better than previous approaches.

By taking advantage of additional organ classes during model training, we obtained a better mean Dice score of 70.3\%. The goal of few-shot segmentation is to construct a \emph{class-agnostic} model by transferring semantic information from the support set to query images for unseen classes. Adding more class types increases training set diversity, prevents the trained model from overfitting to particular training classes, and produces better generalization-capable models. ALPNet~\cite{ouyang2020self} attains this goal with a self-supervised training technique for few-shot segmentation. We note that ALPNet achieves a similar Dice score to our method for some organs, although the experiments reported in~\cite{ouyang2020self} were conducted on different datasets, thus precluding a fair comparison.

\begin{figure*}[ht]
	\centering
	\includegraphics[width=15cm,height=22cm]{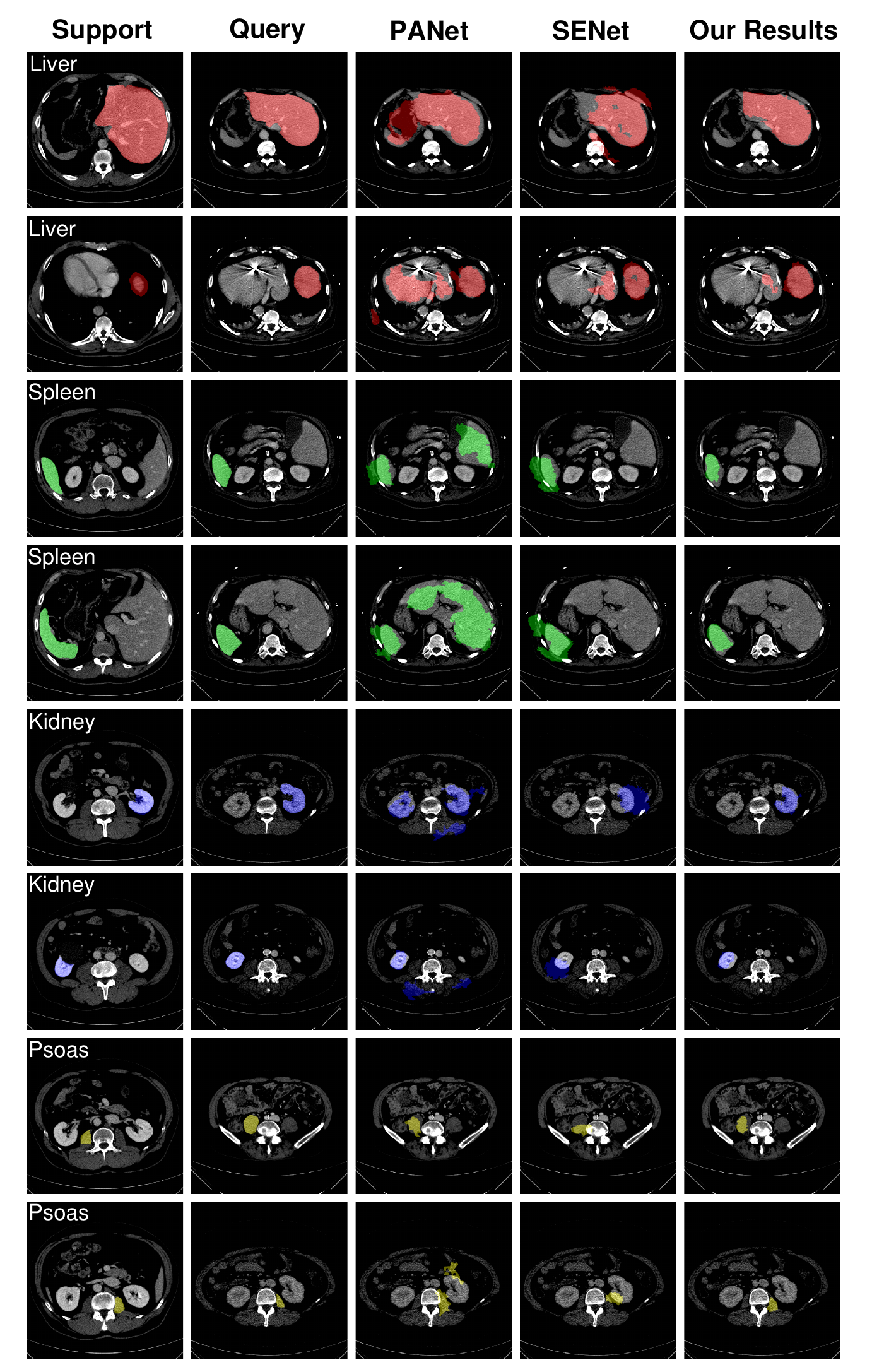}
	\caption{Additional visualizations. (Col.1) Support images and the associated ground-truth masks. (Col.2) Query images and the associated ground-truth masks.
		(Col.3) PANet results. (Col.4) SENet results. (Col.5) Our results.}
	\label{fig:more_vis}
\end{figure*}

Fig.~\ref{fig:visual} presents some segmentation visualizations. Evidently, our method successfully transfers semantic information from support to query images to produce high-quality segmentation results despite different organ sizes and shapes. Fig.~\ref{fig:more_vis} (on the last page of this paper) shows additional segmentation results of our method compared with other approaches. 

\subsection{Ablation Study}
\label{sec:ablation}

\begin{figure}
\centering
\includegraphics[width=0.49\linewidth]{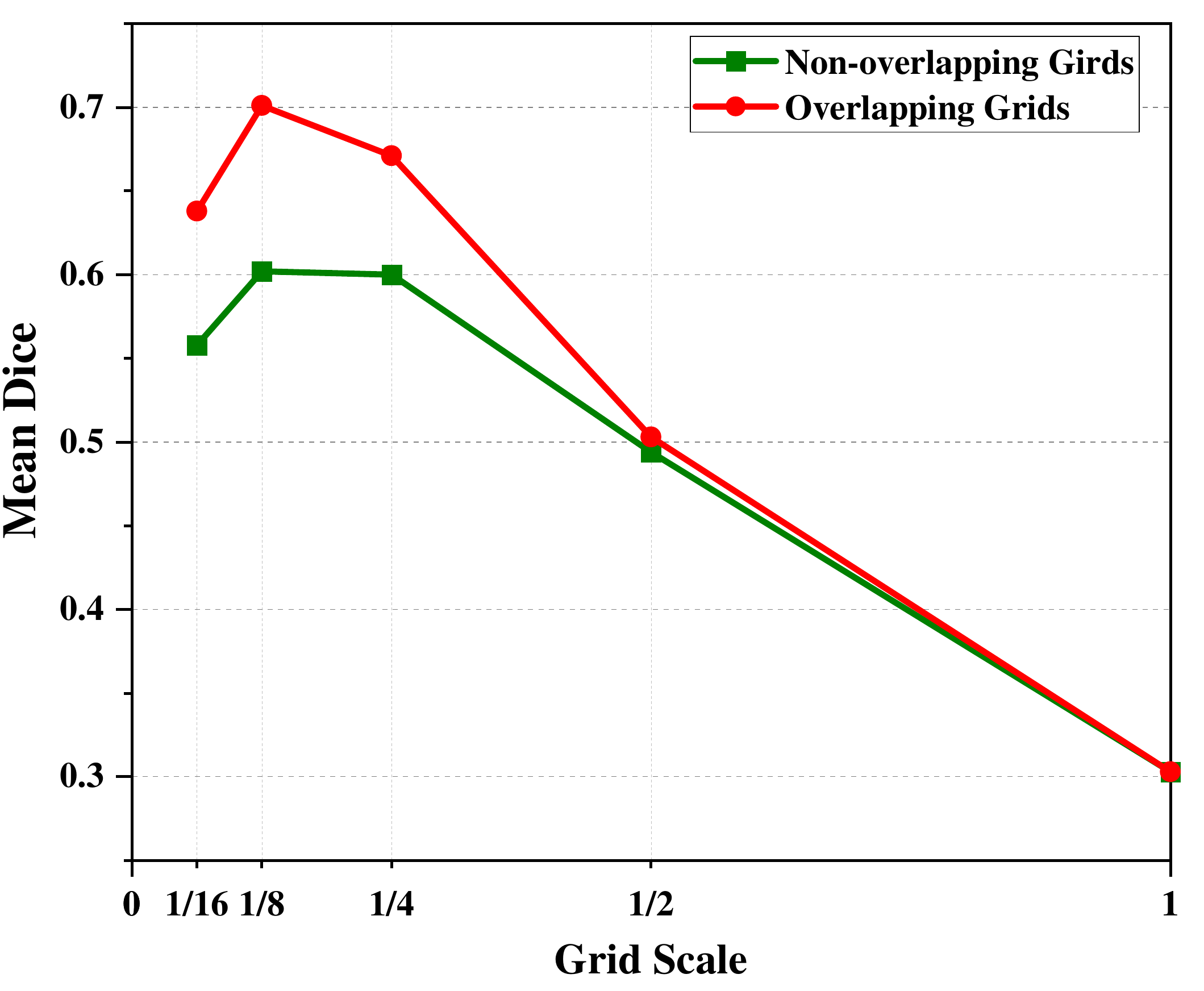} \hfill
\includegraphics[width=0.49\linewidth]{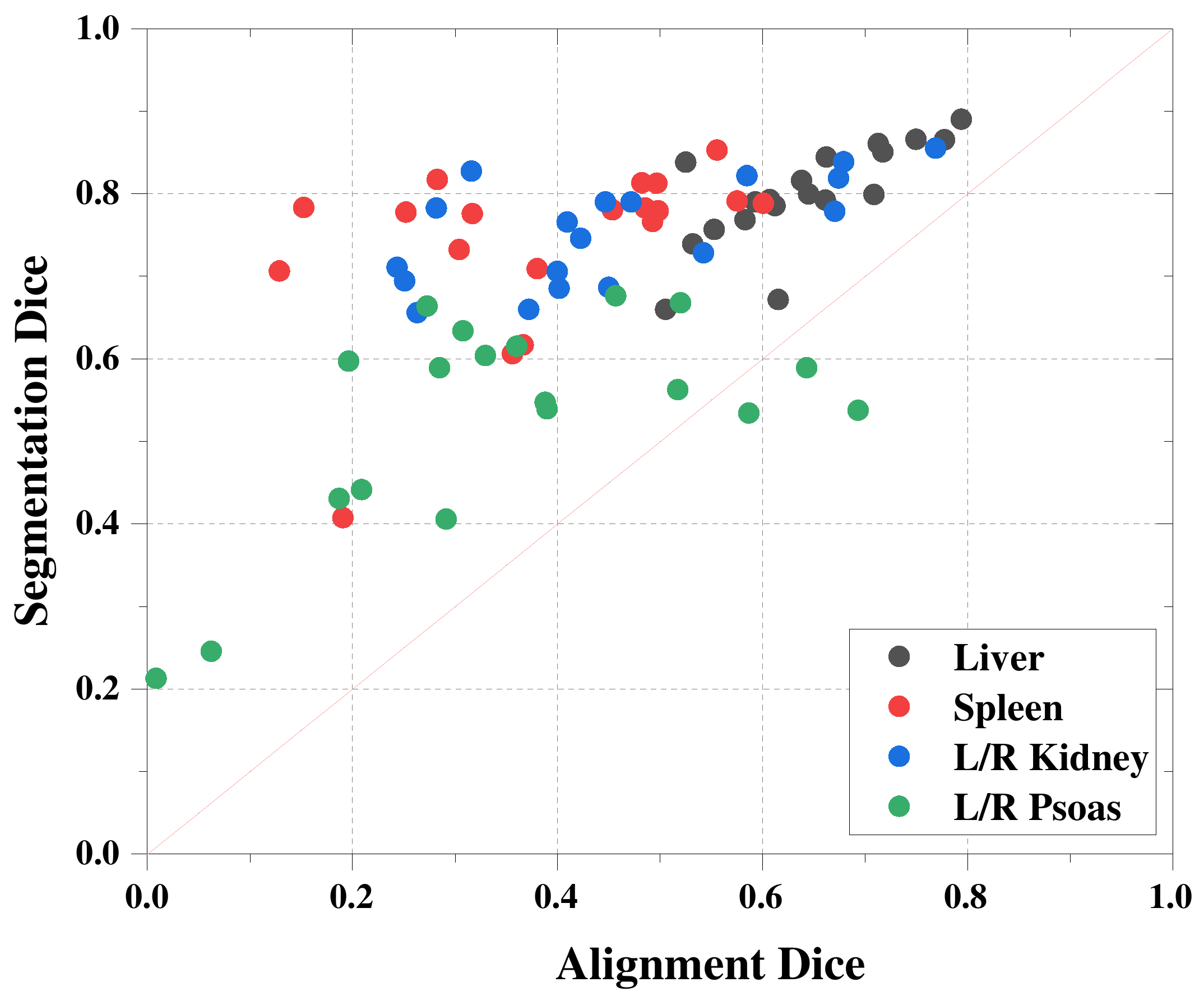}
\caption{Ablation study. (Left) Benefits of selecting an appropriate grid scale and using overlapping image grids. (Right) Performance of our approach for different degrees of spatial misalignment.}
\label{fig:ablation}
\end{figure}

The left graph of Fig.~\ref{fig:ablation} shows the benefit of selecting an appropriate grid scale. When the grid scale $\alpha$ is set to $1/16$ or $1/4$ instead of the optimal value of $1/8$, the location prior may be overly or insufficiently utilized and the mean Dice score drops from 70.3\% to 63.8\% or 67.1\%, respectively. When $\alpha=1$, the grid covers the entire image and the mean Dice score falls to 30.3\%, because the location prior is not utilized at all. The resultant performance drop reveals the significance of location information. The graph also shows that when using non-overlapping rather than overlapping image grids, the mean Dice score decreases from 70.3\% to 60.2\%, thus confirming the benefit of overlapping image grids.

Our approach relies on the assumption that the support and query images have a similar layout and are somewhat spatially aligned. We evaluated its performance when this assumption fails. The right graph of Fig.~\ref{fig:ablation} shows the performance of our approach for different degrees of spatial misalignment. From the testing set, for each support-query pair and each organ class, we calculated the Dice coefficient between the ground-truth support mask and the ground-truth query mask, indicated as ``alignment Dice'' on the horizontal axis of the graph. We also calculated the Dice score between the segmentation result and the ground-truth query mask, indicated as ``segmentation Dice'' on the vertical axis.  The graph shows that our method is quite robust against spatial misalignments---for organs such as \texttt{liver}, \texttt{spleen}, and \texttt{l/r kidney}, many data points on the graph are associated with an alignment Dice of less than 40\%. This implies that for the corresponding support-query pairs, the ground-truth support and query masks have little spatial overlap and, hence, are poorly aligned. Nevertheless, the segmentation Dice scores for these support-query pairs are often higher than 60\%, signifying good performance despite the spatial misalignments between the ground-truth support and query masks. Our method performs less well for \texttt{l/r psoas}, as these are small anatomical structures whose visual appearance is similar to the surrounding structures, making it harder to learn discriminative visual features. Overall this experiment demonstrated that our method has good misalignment tolerance for certain organs.   

\balance

\section{Conclusions}

We proposed a prototype-based method that leverages spatial layout priors in performing few-shot medical image segmentation. Exploiting local prototypes, our novel location-sensitive local prototype network divides the hard problem of segmenting the entire image into easily solvable local region segmentation subproblems, yielding significantly improved performance. In organ segmentation experiments on the VISCERAL dataset, our model outperformed the current state of the art by 10\% in the mean Dice coefficient. Our extensive ablation studies demonstrated the substantial benefits of incorporating spatial information.

\section{COMPLIANCE WITH ETHICAL STANDARDS}

This research study was conducted retrospectively using human subject data of annotated anatomical structures made available in open access by the VISCERAL Consortium. Ethical approval was not required as confirmed by the license attached with the open access data.

\section{ACKNOWLEDGMENTS}

This research was supported by VoxelCloud, Inc. XD and DT are founders of the company.

\small

\bibliographystyle{IEEEbib}
\bibliography{isbi21}

\begin{thebibliography}{10}

\bibitem{ronneberger2015u}
O.~Ronneberger, P.~Fischer, and T.~Brox,
\newblock ``{U-Net}: Convolutional networks for biomedical image
  segmentation,''
\newblock in {\em International Conference on Medical Image Computing and
  Computer-Assisted Intervention}. Springer, 2015, pp. 234--241.

\bibitem{isensee2018nnu}
F.~Isensee, J.~Petersen, A.~Klein, et~al.,
\newblock ``{nnU-Net}: Self-adapting framework for {U-Net}-based medical image
  segmentation,''
\newblock {\em arXiv preprint arXiv:1809.10486}, 2018.

\bibitem{chen2017rethinking}
L.-C. Chen, G.~Papandreou, F.~Schroff, and H.~Adam,
\newblock ``Rethinking atrous convolution for semantic image segmentation,''
\newblock {\em arXiv preprint arXiv:1706.05587}, 2017.

\bibitem{tajbakhsh2019surrogate}
N.~Tajbakhsh, Y.~Hu, J.~Cao, X.~Yan, Y.~Xiao, Y.~Lu, J.~Liang, D.~Terzopoulos,
  and X.~Ding,
\newblock ``Surrogate supervision for medical image analysis: Effective deep
  learning from limited quantities of labeled data,''
\newblock in {\em IEEE International Symposium on Biomedical Imaging}, 2019,
  pp. 1251--1255.

\bibitem{rakelly2018few}
K.~Rakelly, E.~Shelhamer, T.~Darrell, A.A. Efros, and S.~Levine,
\newblock ``Few-shot segmentation propagation with guided networks,''
\newblock {\em arXiv preprint arXiv:1806.07373}, 2018.

\bibitem{DBLP:conf/bmvc/DongX18}
N.~Dong and E.P. Xing,
\newblock ``Few-shot semantic segmentation with prototype learning,''
\newblock in {\em British Machine Vision Conference}. 2018, p.~79, {BMVA}
  Press.

\bibitem{wang2019panet}
K.~Wang, J.H. Liew, Y.~Zou, D.~Zhou, and J.~Feng,
\newblock ``{PANet}: Few-shot image semantic segmentation with prototype
  alignment,''
\newblock in {\em IEEE International Conference on Computer Vision}, 2019, pp.
  9197--9206.

\bibitem{zhang2019canet}
C.~Zhang, G.~Lin, F.~Liu, R.~Yao, and C.~Shen,
\newblock ``{CANet}: Class-agnostic segmentation networks with iterative
  refinement and attentive few-shot learning,''
\newblock in {\em IEEE Conference on Computer Vision and Pattern Recognition},
  2019, pp. 5217--5226.

\bibitem{liu2020prototype}
J.~Liu and Y.~Qin,
\newblock ``Prototype refinement network for few-shot segmentation,''
\newblock {\em arXiv preprint arXiv:2002.03579}, 2020.

\bibitem{roy2020squeeze}
A.G. Roy, S.~Siddiqui, S.~P{\"o}lsterl, N.~Navab, and C.~Wachinger,
\newblock ```{S}queeze \& excite' guided few-shot segmentation of volumetric
  images,''
\newblock {\em Medical Image Analysis}, vol. 59, pp. 101587, 2020.

\bibitem{dang2017learning}
K.~Dang and J.~Yuan,
\newblock ``Learning location constrained pixel classifiers for image
  parsing,''
\newblock {\em Journal of Visual Communication and Image Representation}, vol.
  49, pp. 1--13, 2017.

\bibitem{haghighi2020learning}
F.~Haghighi, M.R. Hosseinzadeh~Taher, Z.~Zhou, M.B. Gotway, and J.~Liang,
\newblock ``Learning semantics-enriched representation via self-discovery,
  self-classification, and self-restoration,''
\newblock in {\em Medical Image Computing and Computer Assisted Intervention}.
  2020, pp. 137--147, Springer.

\bibitem{jimenez2016cloud}
O.~Jimenez-del Toro, H.~M{\"u}ller, M.~Krenn, et~al.,
\newblock ``Cloud-based evaluation of anatomical structure segmentation and
  landmark detection algorithms: {VISCERAL} anatomy benchmarks,''
\newblock {\em IEEE Transactions on Medical Imaging}, vol. 35, no. 11, pp.
  2459--2475, 2016.

\bibitem{DBLP:journals/corr/abs-1810-12241}
A.K. Mondal, J.~Dolz, and C.~Desrosiers,
\newblock ``Few-shot {3D} multi-modal medical image segmentation using
  generative adversarial learning,''
\newblock {\em CoRR}, vol. abs/1810.12241, 2018.

\bibitem{zhao2019data}
A.~Zhao, G.~Balakrishnan, F.~Durand, J.V. Guttag, and A.V. Dalca,
\newblock ``Data augmentation using learned transformations for one-shot
  medical image segmentation,''
\newblock in {\em IEEE Conference on Computer Vision and Pattern Recognition},
  2019, pp. 8543--8553.

\bibitem{ouyang2020self}
C.~Ouyang, C.~Biffi, C.~Chen, T.~Kart, H.~Qiu, and D.~Rueckert,
\newblock ``Self-supervision with superpixels: Training few-shot medical image
  segmentation without annotation,''
\newblock {\em arXiv preprint arXiv:2007.09886}, 2020.

\bibitem{tajbakhsh2020embracing}
N.~Tajbakhsh, L.~Jeyaseelan, Q.~Li, J.N. Chiang, Z.~Wu, and X.~Ding,
\newblock ``Embracing imperfect datasets: A review of deep learning solutions
  for medical image segmentation,''
\newblock {\em Medical Image Analysis}, p. 101693, 2020.

\bibitem{he2016deep}
K.~He, X.~Zhang, S.~Ren, and J.~Sun,
\newblock ``Deep residual learning for image recognition,''
\newblock in {\em IEEE Conf. on Computer Vision and Pattern Recog.}, 2016, pp.
  770--778.

\bibitem{deng2009imagenet}
J.~Deng, W.~Dong, R.~Socher, L.-J. Li, K.~Li, and L.~Fei-Fei,
\newblock ``{ImageNet}: A large-scale hierarchical image database,''
\newblock in {\em IEEE Conference on Computer Vision and Pattern Recognition},
  2009, pp. 248--255.

\bibitem{bottou2010large}
L.~Bottou,
\newblock ``Large-scale machine learning with stochastic gradient descent,''
\newblock in {\em Proceedings of COMPSTAT 2010}, pp. 177--186. Springer, 2010.

\end{thebibliography}

\end{document}